# Real-time Informative Surgical Skill Assessment with Gaussian Process Learning

Yangming Li[1], Randall Bly[2], Sarah Akkina[2], Rajeev C. Saxena[2], Ian Humphreys[2],
Mark Whipple[2], Kris Moe[2], Blake Hannaford[2]

*Abstract*—Endoscopic Sinus and Skull Base Surgeries (ESS-BSs) is a challenging and potentially dangerous surgical procedure, and objective skill assessment is the key components to improve the effectiveness of surgical training, to re-validate surgeons' skills, and to decrease surgical trauma and the complication rate in operating rooms. Because of the complexity of surgical procedures, the variation of operation styles, and the fast development of new surgical skills, the surgical skill assessment remains a challenging problem. This work presents a novel Gaussian Process Learning-based heuristic automatic objective surgical skill assessment method for ESSBSs. Different with classical surgical skill assessment algorithms, the proposed method 1) utilizes the kinematic features in surgical instrument relative movements, instead of using specific surgical tasks or the statistics to assess skills in real-time; 2) provide informative feedback, instead of a summative scores; 3) has the ability to incrementally learn from new data, instead of depending on a fixed dataset. The proposed method projects the instrument movements into the endoscope coordinate to reduce the data dimensionality. It then extracts the kinematic features of the projected data and learns the relationship between surgical skill levels and the features with the Gaussian Process learning technique. The proposed method was verified in full endoscopic skull base and sinus surgeries on cadavers. These surgeries have different pathology, requires different treatment and has different complexities. The experimental results show that the proposed method reaches *100% prediction precision for complete surgical procedures* and *90% precision for real-time prediction assessment*.

## I. Introduction

Endoscopic sinus and skull base surgery (ESSBS) are commonly performed operations of the paranasal sinuses and skull base for reasons such as neoplasm, cancer, and chronic rhinosinusitis (CRS) [1]. The endoscopic surgical techniques have decreased surgical morbidity and postoperative recovery time in over 350,000 Endoscopic Sinus and Skull Base Surgery (ESSBS) annually [1]. Despite the benefits from these techniques, the surgical procedures are made more challenging and potentially dangerous, due to the limited field of view, the narrow anatomic confines and the proximity to critical neurovascular structures [2, 3]. Because of the complexity of ESSBS procedures, the ESSBS skill assessment is one of the most needed and challenging task [4].

Surgical skill assessment serves as the cornerstone in the model residency training system introduced by Sir William Halsted[5]. In this apprentice-style system, trainees heavily rely on assessment for efficient dexterity improvement. The early skill assessment was performed by expert surgeons watching the trainees' performance [6]. Later, this type of assessment was advanced to provide more objective assessments, by evaluating according to global and task-specific checklists [7]. These assessment methods are effective but not efficient, as a significant amount of time from expert surgeons is required. The need for automatic objective surgical skill assessments is becoming unprecedented important in surgical training, as the Accreditation Council for Graduate Medical Education(ACGME) requires continuously assessing all medical residents by competency-based measures [8] [9].

Surgical skill assessment is also increasingly demanded for demonstrating (attending) surgeons' skills. The Agency for Healthcare Research and Quality reported more than 32,000 surgery-related deaths in 2000, which also leads to 9 billion dollars cost and 2.4 million extra hospital days [10]. Coincidental, a study shows only 34% of surgical trainees felt that they were receiving adequate training [11]. Pressure and need on demonstrating surgeons are skillful and are capable of maintaining the skill levels are increasing [12].

Surgical skill assessment is also needed to be applied to real-time skill evaluation in operating rooms. It is well known that surgeons' performance not only depends on their skillfulness but also varies among cases. Evaluating cases in operating rooms can assist surgeons in ensuring surgical performance, improving surgical outcomes and decreasing revision surgery rates.

Because of the complexity of real surgical procedures, most of the existing assessment studies can only be applied to the medical training. These studies use video and motion or both as the source data. Video data contains richer information, but it is challenging to precisely and reliably recovery instrument position and segment into surgical tasks, thus is commonly used for coarse-grained surgical activity pattern study, such as surgical work-flow study [6]. Motion data is used for fine-grained surgical pattern study. Early studies demonstrated that for well-defined surgical tasks, for example, suturing and knot-tying, simple statistical metrics, such as completion time, total movement, and motion economy, can be powerful for skill assessment [13]. More recent research adopted Markov models(MM) or conditional random fields(CRF) for seeking finer metrics, either using motion data along [14] or combing motion data with force or tissue contacting information [15]. In these methods, once gestures and surgical status are defined as the phonemes defined in natural language processing, applying MM and CRF on the problem becomes straightforward. These techniques can also be extended to operating rooms, if surg-eries are highly structured, such as cholecystectomy [16]. For unstructured surgeries, if informative features can be found, it is still possible to access surgical skills with operating room data. For example, Poddar et al. successfully assess skill using motion patterns of brushing activities in septoplasty and the average accuracy reaches 72% [17].

In this work, we present a novel Gaussian Process based-learning method that can be applied to surgical training, as well as demonstrating attending surgeons' skills and real-time operating room operation assessment. Being different from existing assessment algorithms, the proposed method assesses skills based on real-time kinematic features, has no dependency on specific surgical tasks, and has no assumption on surgical procedures. To be more specific, the central contributions of this paper are:

- We propose a Gaussian Process-based Learning for automatic objective surgical skill assessment method. The method incrementally learn optimal motion patterns, provide real-time informative assessment results, and can be used for surgical robot motion planning.
- The proposed method has no assumption on the types of surgical tasks and is independent of the statistics, such as the length of surgery, motion economy, etc. Thus can be applied to part of surgeries and be used for real-time skill assessment in operating rooms.
- The proposed method breaks the bottleneck of using data-driven learning algorithms in endoscopic surgeries. Traditionally, assessing endoscopic surgeries utilizes the complete surgeries, thus it is too expensive to achieve sufficient data for model learning. The proposed method uses surgical instrument relative motion to reduce the dimensionality, and uses motion data kinematic features for learning, thus one complete surgery can be used as thousands of dataset for training.
- The proposed method utilizes widely available surgical navigation systems, and has no impact on existing surgical protocols, thus potentially can be applied to all ESSBSs.
- We demonstrated the effectiveness and the efficiency of the proposed method on real ESSBS on cadavers.

The rest of the paper is organized as follows. The next section presents the preliminaries, including the related works and the motivation and the rationale of this work. Section III presents the proposed method in detail. Section IV demonstrates the application of the proposed method on ESSBSs on cadavers. Conclusions are drawn in the last section.

## II. PRELIMINARIES

### A. Related Works

The advance of surgical technology increases the complexity of surgical procedures. These changes not only make the surgical training facing unprecedented challenges of teaching more surgical skills in a shorter period of time but also bring in the need for self-evaluation and operating room operation monitoring, as attending surgeons are thrived to improve surgical outcomes and embrace new surgical techniques.

Although both trainees and experienced surgeons need surgical skill assessment technology, almost all existing methods apply to the surgical training procedure [6, 9]. Existing surgical skill assessment methods can be generally categorized into three groups: human grading, descriptive statistics, statistical language models of surgical motion. They often utilize two different types of source data: surgical scene (direct observation, surgical video), and surgical motion (instrument motion, surgeon hand motion, surgeon eye movements).

Human grading directly evaluates surgeries (mainly) based on videos and direct observations. Because of the insight that teaching dexterous skill relies on the apprentice-style training, this observational approach (OA) was the first adopted method and is still widely used [7]. The structured human grading (SHG) was advanced based on OA by introducing global and standard checklists, which provides more objective feedback [7]. Because these methods require a significant amount of time from expert surgeons to review the surgical procedures, crowdsourcing was introduced to ease the burden of experts by replacing expert with a large amount of nonmedically trained people [18].

Descriptive statistical methods use statistical features to differentiate surgeon dexterity. These methods often utilize statistical metrics, such as motion economy, peak forces, torques, tissue damage, motion repeatability, and path following, thus often are valid for specific surgical tasks [12, 19–21].

Descriptive language models also utilize statistics. Because the surgical procedures have intrinsically similarity with languages as they are both sequential, methods similar to language processing are developed to analyze surgical skills [14, 22]. These methods build representative models based on experts and the comparative similarity to the models are served as the skill assessment results. As real surgical procedures vary with respect to patients and pathology, it is generally difficult to extend these methods for real surgery assessment[9].

### B. Motivation and Rational

From the discussion in Subsection II-A it is clear that existing automatic objective surgical skill assessment methods 1) are often designed for specific surgical tasks or procedures for the training purpose; 2) often depend on complete surgical procedures and cannot assess in real-time [6, 9]. Some of these methods also depend on specific instruments and can not be directly extended to surgeries in operating room [6]. As the results, to our best knowledge, none of these methods are designed for attending surgeon self-evaluation in operating rooms, or intraoperative real-time surgery monitoring.

This paper aims to meet the need for real-time OR applicable automatic objective surgical assessment methods. The proposed method aims to provide informative feedback and real-time monitoring to surgeries in OR. Based on our observations of ESSBS surgeries in operating rooms, this work proposes:

- Using real-time observations of surgical motion for surgical skill assessment. This is inspired by the fact that experienced surgeons can assess surgical skills after a short period of observation, instead of watching the complete surgeries.
- Using kinematic features, instead of the sequence of operations, for the assessment. This is inspired by the observations that 1) attending surgeons often correct trainees motion patterns, such as speed, angle, etc., 2) non-medically trained personnel has no knowledge on surgical procedures but can assess skills in crowd-sourcing methods.
- Using the relative motion between surgical instruments and endoscope for assessment. In Image Guided Surgery (IGS), experiences surgeons perform surgeries based on endoscopic videos. In these videos, the endoscope serves as the global reference and both instruments and anatomical regions moves with respect to the endoscope, therefore, in IGS, the relative movement not only reflects the dexterity but also shows the hand-eye coordination and the hand-hand coordination.
- Using the relative motion can reduce the feature space and decrease the difficulty of finding the features for skill assessment.

## III. RELATIVE MOTION BASED GAUSSIAN PROCESS LEARNING FOR AUTOMATIC OBJECTIVE SURGICAL SKILL ASSESSMENT

### A. Data Acquisition

In order to introduce the surgical skill assessment into operating rooms, this paper adopts FDA approved commercial surgical navigation systems for data collection[23–26]. Surgical navigation systems are widely used in endoscopic surgeries, especially the endoscopic skull base and sinus surgeries. This technology establishes a global tracking system with cameras, infrared sensors or electromagnetic emitters. Through simultaneously tracking the patient and the surgical instruments, the relative position and movements between the patient and the instruments can be reliably calculated.

In this work, the proposed method adopts commercial surgical navigation systems to demonstrate the maximized applicability. Medtronic (Minneapolis, MN, USA 55432) StealthStation® S7® surgical navigation system was used in this work for data acquisition. Although the proposed method has no impact on existing surgical procedures and can be safely applied in OR, we only verified the proposed methods in cadaver experiments because of the restriction of Medtronic research licensing of the surgical navigation system. The data acquisition system is shown in Fig. 1, in which Fig. 1a shows Medtronic S7® and 1b shows the tracked endoscope and the microdebrider.

### B. Data Preprocess

Medtronic S7® is an infrared optical global tracking based navigation system. In this system, the cameras directly observe the trackers' positions, and through calibrating the relative position between the trackers and the instruments' tips, the movements of instruments are known to the tracking system.

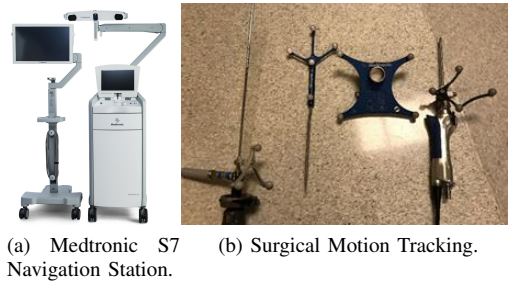

(a) Medtronic S7 Navigation Station.  (b) Surgical Motion Tracking.

Fig. 1: Data Collection.

This type of tracking systems has reliable tracking performance and is widely used in ESSBSs.

As the motion relative to the endoscope is used for surgical skill assessment, all motions are first projected to the endoscope coordinate. We defined the endoscopic coordinate as: *the origin is the center of the endoscope tip, the X-positive is the vector starting from the origin and pointing to the hand-held end of the endoscope, and the X-Z plane is formed by the X-axis and the Cranial vector* (point up to the tip of the head). This coordinate definition was inspired by the traditional visualization of endoscopic view on flat monitors, where "up" points to the Cranial vector and the "depth" direction is parallel with the endoscope. Notice that if angled endoscopes are used in surgery, the definition needs to be adjusted to reflect the change.

Although the commercial surgical navigation systems track both the endoscope and the instrument, further processing is required to improve the reliability of the instrument position with respect to the defined endoscopic coordinate. This is because: 1, the commercial surgical navigation systems have about 1 mm tracking precision, but the errors are amplified in the projection; 2, the tracking systems have low and varying sampling rate and high occlusion rate (see Tab.I); 3) the tracking systems only provide the tracker position and orientation information, and velocities are not available. We addressed these limitations by closing the tracking loop with detecting instruments' positions from endoscopic images, based on a new deep network, called ToolNet ([27]).

The refined instrument positions are ready for feature extraction. Inspired by observations of ESSBSs, the following features are the candidates for learning the skill model from motion data, including:

- the 3-dimensional instrument tip position in the endoscopic coordinate;
- the distance between the instrument tip and the endoscopic tip;
- the distance from the instrument tips to the endoscope line;
- the instrument tip speed in the endoscopic coordinate;
- the angle between the instrument and the endoscope;
- the angular speed of the instrument with respect to the endoscope.

Given the definition of the endoscope coordinate and the relative positions of the endoscope and the instrument, it is nearly trivial to calculate most of the features, except the speed.

As the asynchronized data retrieving was adopted for extracting the endoscope positions and the instrument positions from Medtronic S7®, in order to minimize the sampling time difference between the motions, the sampling rate of positions are below 20 Hz and varies. The extended Kalman filter under the constant speed model was adopted to smooth the speed estimation from the position observation [28]. This is important to the proposed method as it is essentially an interpolation method, as explained in the next subsection.

### C. Gaussian Process Classification for Automatic Objective Surgical Skill Assessment

It is challenging to automatically objectively assess surgical skill because it is difficult to find effective features that reliably predict skill levels. Because of the difficulty and the cost of achieving surgical data, it is difficult to directly utilize the state-of-art deep learning research. One of the contributions of this work is that we propose to assess surgical skills on data points, instead of complete surgeries. This grows the scale of datasets from dozens to hundreds of thousands, but it is still far away from millions. However, latest researches on neural networks based expert systems [29, 30], learning-based diagnosis [31–33] and neural networks based motion pattern processing [34] inspired us to explore the learning methods that are effective on smaller datasets.

Gaussian Process Classification is adopted to learn the skill assessment from the features, in another word, it learns the model of the kinematic features with respect to the skill level. We adopte this model inspired by the researches that demonstrated the GP's superior capability on learning strongly nonlinear dynamic models [35]. Gaussian Process Classification is a form of supervised learning [36]. It is similar to Bayesian Linear Regression(BLR) (see chapter 3.3 in [37]), in the sense that, given training set $(X,Y)$, they both probabilistically estimate the expected output $Y^\star$ with respect to the input $X^\star$, under the Bayesian framework: $p(Y^\star|X^\star,X,Y)$. To be more specific, under the framework, if the relationship between the input $X$ and output $Y$ is denoted as: $Y = f(X) + \omega$, where $\omega \sim \mathcal{N}(0, \sigma_n^2)$, then according to the Bayesian rule:

$$p(Y^\star|X^\star,X,Y) = \int p(Y^\star|X^\star,f)p(f|X,Y)df. \quad (1)$$

For the given prior of the model: $p(f)$, the posterior of $f$ can be expressed as:

$$p(f|Y,X) = \frac{p(Y|X,f)p(f)}{p(Y|X)}. \quad (2)$$

A simple example is the linear model $f(X) = X^T w$. Under the Bayesian framework, the likelihood $p(Y|X,f)$ can be inferred from the training data as: $Y|X,w \sim \mathcal{N}(X^T w, \sigma_n^2 I)$, where $I$ was an identity matrix, $X$ is a $n \times k$ design matrix and $w$ is a $k \times 1$ weight vector [37].

Gaussian Process Classification can be considered as binarized Gaussian Process Regression (GPR). GPR adopts the same Bayesian framework but utilizes the kernel based schemes, thus it has broader applicability and allows the data "speak" for themselves. The reason that kernels can be utilized in GPR is the method is built on the concept of Gaussian process. A Gaussian process is a collection of random variables and any finite number of which follows a joint Gaussian distribution [36]. Under the Gaussian Process, the joint distribution of the training data $Y$ and the prediction result $Y^\star$ is:

$$\begin{bmatrix} Y \\ Y^\star \end{bmatrix} \sim \mathcal{N}\left(0, \begin{bmatrix} K & K^{\star T} \\ K^\star & K^{\star\star} \end{bmatrix}\right) \quad (3)$$

, where

$$K = \begin{pmatrix} k(X_1,X_1) & \cdots & k(X_1,X_n) \\ \vdots & \ddots & \vdots \\ k(X_n,X_1) & \cdots & k(X_n,X_n) \end{pmatrix} \quad (4)$$

is learned from the training data, and $K^\star = (k(X^\star,X_1) \cdots k(X^\star,X_n))$ and $K^{\star\star} = k(X^\star,X^\star)$ are calculated according to the selected Gaussian Kernel. Therefore, the prediction $Y^\star$ given the known data $Y$ is

$$Y^\star|Y \sim \mathcal{N}(K^\star K^{-1} Y, K^{\star\star} - K^\star K^{-1} K^{\star T}) \quad (5)$$

.

## IV. EXPERIMENTAL RESULTS AND DISCUSSION

### A. Experimental Setup

The proposed method has no impact on existing surgical protocols and can be directly applied to operating room surgeries. However, in this work, the verification of the method was conducted on cadaver experiments (as explained in Fig. 2), *due to the licensing of our Medtronic S7 surgical navigation system*. Eight surgeons (one 2nd year, two 3rd year and one 4th year residents, and four attending surgeons) were instructed to perform common Otolaryngology procedures on four cadaver heads. Each of the eight surgeons performed on one side of the cadaver so the surgical procedures do not interfere with each other. The surgeons performed on maxillary antrostomy, and anterior and posterior ethmoidectomy, but not sphenoidotomy, to represent the most commonly performed procedures (approximately 300,000 cases performed yearly in the United States [38]). During the procedure, surgeons freely select proper surgical operations as they needed, and the operations include medializing the middle turbinate, removing uncinate process, maxillary antrostomy with bone and soft tissue removal, ethmoid bulla entering, and anterior and posterior ethmoidectomy with the removal of bone and soft tissue. Surgeons used a 4 mm zero-degree endoscope and a straight 4 mm microdebrider (Straightshot M4, Medtronic plc, Minneapolis, MN, USA 55432).

### B. Feature Extraction

Features are extracted from the pre-processed (precision is refined by ToolNet [27]) projected motion data. For the position related features, we adopt the asynchronized method to localize the endoscope and the microdebriter with the surgical navigation system, in order to ensure that the endoscope and the microdebriter are localized at the nearly same time. We do not further smooth the trajectories and resample the positions.

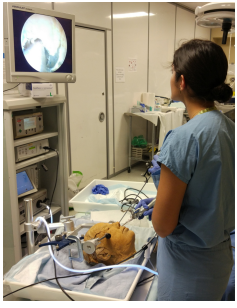

Fig. 2: Cadaver Experiments.

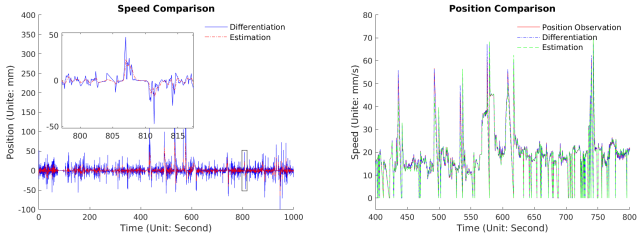

(a) Speed Estimation Comparison.   (b) Speed Estimation Verification.

Fig. 3: Estimating Speed from Positional Observations.

For speed-related features, we smooth the estimations, because: 1) the sampling rates are low and widely vary; 2) it has missing data dues to occlusion (Table I). Because the surgical navigation system is essentially a global tracking system, and the localization errors do not accumulate with time, the Kalman filter was adopted for estimation. Given the human motion characteristics, the constant velocity model was used for the smoothing. The standard deviation for the positional observations and the speed are 0.3$mm$ and 1$mm/s$, which were empirically selected given the Medtronic S7 system specifications. The estimation results are visualized in Fig. 3. The results are verified by comparing the positional observations, with the integration of the speed estimation and integration of the speed differentiated from the raw positional observations.

TABLE I: Statistics of Motion Data from Commercial Surgical Navigation System.

|    | Sampling Rate High (Unit: Hz) | Sampling Rate Low (Unit: Hz) | Tracking Rate (Unit: %) | Duration (Unit: s) |
|----|---|---|---|---|
| E1 | 14.71 | 6.62 | 82.38 | 1005.80 |
| E2 | 12.66 | 6.67 | 96.11 | 543.99 |
| E3 | 13.51 | 6.49 | 59.23 | 985.24 |
| E4 | 15.63 | 6.14 | 94.14 | 297.62 |
| N1 | 14.71 | 6.25 | 76.47 | 1093.03 |
| N2 | 12.50 | 6.25 | 74.84 | 1385.20 |
| N3 | 14.92 | 6.41 | 88.49 | 810.31 |
| N4 | 14.92 | 6.33 | 97.33 | 558.76 |

E: Expert; N: Novice.

## C. Feature Selection

The proposed method utilizes Gaussian Process to "learn" the relationship between the features and surgical skill levels. To be more specific, the proposed method differentiates skill levels in the feature space, through Gaussian Process Classification. Therefore, effective features increase the differentiability in the feature space, while the ineffective features increase the feature space ambiguity. As the feature candidates are inspired by surgery domain knowledge and the observations to surgery, we further directly "observe" feature distributions to refine the feature selection.

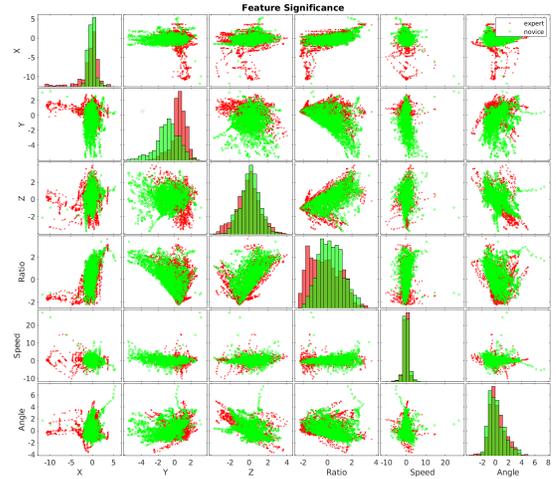

Fig. 4: Feature Significance

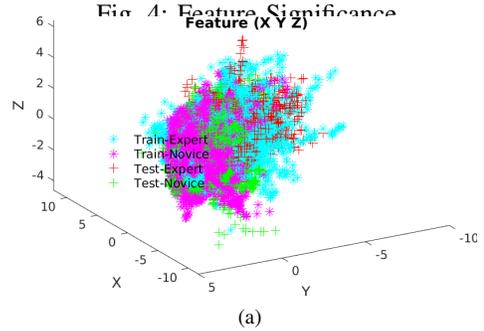

(a)

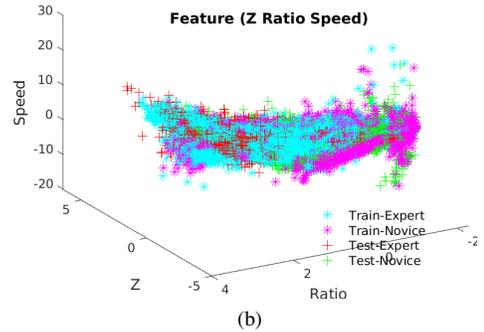

(b)

Fig. 5: Feature Spatial Distribution.

Fig. 4 visualizes the significance of the features. Fig. 5 visualizes the features' geometrical distributions for verification. From the figures it is clear that among all the feature candidates, the projected relative positions (X, Y, and Z), the distance ratio, and the speed have better differentiability for the skill levels, thus are selected in the proposed method.

## D. Skill Differentiation with Gaussian Process Learning

The Gaussian Process-based classification learns to assess surgical skill levels based on the selected features. In order to fully demonstrate the effectiveness of the proposed method, *40% of the two cadaver experiments data is randomly selected for training*, and the rest of the data is reserved for testing.

As one of the major advantages of the proposed method, it assess the surgical skills and produces the confidence of the assessment. This is an exciting feature because it matches our understanding of kinematic features based surgical skill assessment, which is all motions indicate skill levels, but some of them are strong indicators than the others. Intuitively, the assessment results with high confidence are associated with the "signature" motion patterns that strongly indicate the skill level, while the low confidence ones are associated with motion patterns that are similar between experts and novices.

Fig. 6 shows the performance of the proposed method. The confidence levels (inverse of estimation uncertainty) have significance impact on the prediction results. In the figure, the X-axis indicates the estimation uncertainty, where 0 means absolutely sure and 1 means not sure at all; the Y-axis is the ratio of the assessment with respect to the total data point. The red colored solid line shows the ratio of wrong predictions. With the decrease of the confidence level, the proposed method becomes more prudent in making assessment, and the prediction precision increases. The dot-dashed green line shows the false positive rate, which is defined as predicting a novice as an expert. The dashed blue line shows the false negative rate, which is defined as predicting an expert as a novice. The false positives and the false negatives both increase with the decrease of the confidence level, but we can clearly see that false positive rate is lower than the false negative, which indicates that the proposed method is more sensitive to expert motions. This might suggest that expert surgeons have strong motion patterns. Naturally, with the increase of the confidence level, the method is more prudent and make less assessment, as indicated by the cyan colored star-dashed line.

The proposed method can produce summative assessment method, in order to compare with classical summative skill assessment method. To do so, we simply use the weighted summation of real-time assessment results, where the results are 1 for experts and -1 for novices, and the weight is the assessment confidence. If the summative result is bigger or equal than 0, it is expert, otherwise, it is novice. The proposed method reaches an impressive *100% assessment precision* in our experiments.

It is also straightforward to provide informative feedback, because the assessment is performed in a real-time manner and all related information are already available. Software has been developed to provide these information in a easy-to-understand way. While surgeons are performing surgeries, the skill score and the corresponding confidence are shown in real-time. While a low-skill-level movement is detected, the geometrical distributions between the kinematic features from the movement are compared with the distributions from high-skill-level movements, and the corresponding endoscopic video segment is also shown with the comparison because the video also helps surgeons to understand the problem.

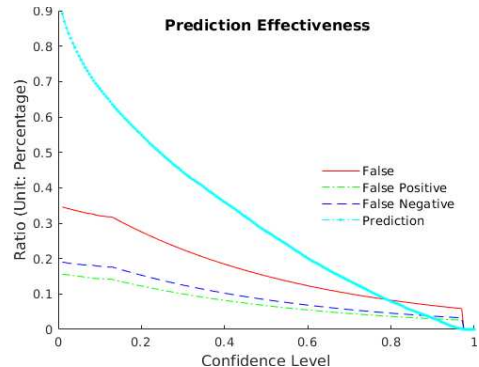

(a) Prediction Rate with respect to the Confidence Level.

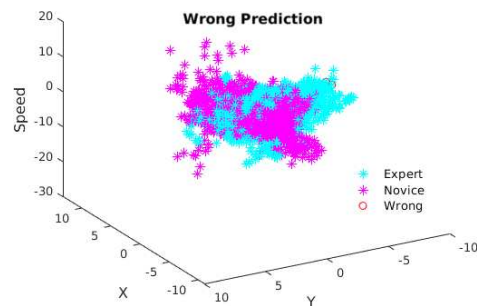

(b) Wrong Predictions.

Fig. 6: Surgical Skill Assessment Effectivenss.

## V. CONCLUSION

Research on surgical skill assessment was originally driven by the fact that surgical skill assessment is a key component in the modern apprentice-style surgical training system. However, with the fast evolution of modern endoscopic surgical technology, the surgical procedures are increasingly challenging, and automatic objective surgical skill assessment is unprecedentedly needed for not only surgical trainee education, but also attending surgeons demonstrating their competence and improving operating room surgical outcomes.

This work proposes a novel relative motion based Gaussian Process learning method for real-time automatic objective surgical skill assessment in ESSBSs. Comparing with existing methods, the one proposed in this work can be generally applied to various surgical procedures in ESSBSs, is independent of the length of surgical procedure, and makes estimation in real-time. More importantly, the proposed method only uses surgical motion data, which is currently available in almost all ESSBSs in US. Real surgeries on cadavers verify that the proposed method reaches 90% prediction precision in real-time, and 100% prediction precision summatively.